\documentclass[journal]{IEEEtran}

\usepackage{graphicx} 
\usepackage{epstopdf} 
\usepackage{subfigure}
\usepackage{caption}
\usepackage{adjustbox} 
\usepackage{multirow} 

\usepackage{pseudo}

\usepackage{amsmath, amsfonts,amssymb,bm}

\usepackage[T1]{fontenc}

\usepackage{multirow}

\usepackage{array}

\usepackage{url}
\usepackage{breakurl}

\def\naive{{na\"{i}ve}}

\def \B0{{\mathbf 0}}

\def \B0{{\mathbf 0}}

\def \w{{\mathbf w}}
\def \Br{{\mathbf r}} 
\def \s{{\mathbf s}}
\def \q{{\mathbf q}}

\def \d{{\mathbf d}}

\def \u{{\mathbf u}}

\def \be{\begin{equation}}
\def \ee{\end{equation}}

\begin{document}

\title{Personalized and situation-aware multimodal route recommendations: the FAVOUR algorithm} %

\author{
	\IEEEauthorblockN{Paolo~Campigotto\IEEEauthorrefmark{1}, Christian~Rudloff\IEEEauthorrefmark{2}, Maximilian~Leodolter\IEEEauthorrefmark{2} and Dietmar~Bauer\IEEEauthorrefmark{3}}
	\\
	\IEEEauthorblockA{\IEEEauthorrefmark{1} TU Dortmund
	\\ Paolo.Campigotto@tu-dortmund.de}
	\\
	\IEEEauthorblockA{\IEEEauthorrefmark{2} Austrian Institute of Technology AIT
	\\ \{Christian.Rudloff,Maximilian.Leodolter\}@ait.ac.at}
	\\
	\IEEEauthorblockA{\IEEEauthorrefmark{3} University Bielefeld
	\\ Dietmar.Bauer@uni-bielefeld.de}
\thanks{Most of this work has been done while all authors were with the Mobility Department, Dynamic Transportation Systems, AIT Austrian Institute of Technology GmbH, Giefinggasse 2, 1210 Vienna, Austria. 
}

}

\maketitle

\begin{abstract}
Route choice in multimodal networks shows a considerable variation between different 
individuals as well as the current situational context. Personalization and situation awareness of recommendation algorithms are already common in many areas, e.g., online retail. However, most online routing applications still provide shortest distance or shortest 
travel-time routes only, neglecting individual preferences as well as the current situation. Both 
aspects are of particular importance in a multimodal setting as attractivity of some transportation modes 
such as biking crucially depends on personal characteristics and exogenous factors like the weather.

As an alternative this paper introduces the FAVourite rOUte Recommendation (FAVOUR) approach to provide personalized, situation-aware route proposals
based on three steps: first, at the initialization stage, the user provides 
limited information (home location, work place, mobility options, sociodemographics)
used to select one out of a small number of 
initial
profiles. Second, 
based on this information, a stated preference survey  
is designed in order to sharpen the profile. 
In this step a mass preference prior is used to encode the prior knowledge on preferences 
from the class identified in step one. 
And third, subsequently the profile is continuously updated during usage of the routing services. 
The last two steps use Bayesian learning techniques in order to incorporate information 
from all contributing individuals.

The FAVOUR approach is presented in detail and tested on a small number of survey participants. 
The experimental results on this real-world dataset show that FAVOUR generates better-quality recommendations w.r.t. alternative learning algorithms from the literature. 
In particular the definition of the mass preference prior for initialization of step two 
is shown to provide better predictions than a number of alternatives from the literature. 
\end{abstract}

\IEEEpeerreviewmaketitle

\section{Introduction}
\label{sec:intro}

\IEEEPARstart{W}{hile} personalized, situation-aware recommendations are nowadays widespread in
online retailing, online routing tools still predominantly provide the shortest or the 
fastest route only, neglecting personal preferences and varying situations. However, personalization 
and situation awareness is of particular importance in multimodal route choice as some 
options might not be available all the time (cars are often shared between family members)
or attractive (biking in rain is much less attractive than in bright sunshine). 

Modern information technology provides means for collecting personal information via 
cookies in browsers or app usage on personal smart-phones. Furthermore, information provision is exploited and accepted by users by the very same channels. Correspondingly the provision of personal, situation-aware route proposals appears to be technically feasible and socially accepted. 

At the same time users might not want to personalize the suggestions themselves, and, in many cases, would not be able to, as their preferences are not explicit. However, the preferences of the users can be inferred from their revealed choices. 
This motivates the development of recommendation algorithms that learn the users' preferences from their past choices and afterwards, based on the learned information, propose the most adequate route (including the transportation modes) for a new trip 
(see e.g.~\cite{personalizedITS_2013,Nuzzolo2015a}). 

The adoption of new technology is very sensitive to the perceived accuracy of the first 
recommendations. A new service that does not perform well on the first occasions a user 
tests the system, will not succeed in being used long enough in order to learn from the 
preferences of the users. On the other hand, the users also do not want to fill
in long questionnaires before starting to use the system.
Thus it is crucial to strike a balance between accuracy from the 
outset and the burden of initializing the system. 

In this paper such a method, in the following referred to as 
the FAVourite rOUte Recommendation (FAVOUR) algorithm, is proposed. It is based on three 
stages, where each stage sharpens the user profile providing information on the users' preferences. 
In the {\bf first stage} a few preliminary questions at the time of registration to the service are 
asked. 
These questions 
include  possible 
sociodemographic information (age, profession, gender, ...), transportation options, home location and 
workplace location or other significant places (location of school or grocery shop) such that at {\bf stage two}  
routing decisions in the 
form of binary comparisons (route A or route B?)  can be provided 
that the user relates to in artificial but realistic stated-preference-off-revealed-preference~\cite{Train_Wilson_08}  type questions. 
This helps the respondents in comprehending the presented information better and thus in ultimately making more informed decisions (see, e.g.,~\cite{rose_designing_2008} or~\cite{hensher_how_2006}).

{\bf Stage three} denotes the usage of the personalized routing algorithm. Routes are proposed based on the estimated personal profiles of the users
and feedback on the actually chosen alternative is gathered in order to further sharpen the 
profile. At this stage extensive use is made of Bayesian learning methods. 

This paper focuses on stage two of the FAVOUR algorithm. 
In particular the paper argues that Bayesian learning is beneficial  
in order to obtain good route recommendations already based on a small number of test questions. 
The key in this direction 
will be the generation of {\em mass preference priors} that encode ``typical'' preferences that are subsequently adapted to the user and the situation by Bayesian learning strategies. The main contribution of the paper in this respect are:
\begin{itemize}
	\item the proposal of an overall framework for devising a route recommendation system adapting to the stated and revealed preferences of its users including initialization procedures;
	\item the proposal of a particular transfer learning~\cite{transferLearning10} strategy in order to obtain good predictions based on only a small number of test questions;
	\item the investigation of the achievable accuracy as a function of the number of test questions asked.
\end{itemize}
It is shown in the paper that our approach provides superior performance for a small number of test questions asked in comparison to non-Bayesian techniques using mixed logit models as well as in comparison to state-of-the-art Bayesian methods using alternative initialization methods.

The rest of this paper is structured as follows.
First, the literature related to FAVOUR is discussed. The problem setting is defined in Sec.~\ref{sec:ps}. Sec.~\ref{sec:FAVOUR} and Sec.~\ref{sec:cold-start} detail the FAVOUR algorithm and 
motivate its design choices. This is followed by the experimental evaluation of FAVOUR and the 
comparison with two benchmarking algorithms from the literature:
the likelihood-maximization approach from~\cite{Eiter2014} and the state-of-the-art personalized route-recommendation technique introduced in~\cite{personalizedITS_2013}.
Finally, Sec.~\ref{sec:conclusion} draws conclusions and highlights potential research directions.

\section{Related Literature}

The quest for personalized routing systems based on traveler preferences has been 
highlighted in the transportation system literature~\cite{Nuzzolo2015a,personalizedITS2012,chorus_personal_2011,VLDB15,ICDE15}. 
In~\cite{chorus_personal_2011} a willingness-to-pay model is estimated. While car 
drivers are not receiving a high utility from being presented transit information, the willingness 
to pay for personalized information that is relevant for the current travel-situation %
is notable. Personalized intelligent transportation systems (ITS) are therefore worthwhile. 

Methods generating personalized-route recommendations for drivers are presented also in~\cite{park_adaptive_2007,VLDB15,ICDE15}. Like FAVOUR, they assume that humans take traveling decisions based on several (conflicting) criteria, and that different users trade off these criteria differently (e.g., time-efficient driving 
versus fuel-efficient driving). Unlike FAVOUR,
papers~\cite{VLDB15,ICDE15} do not consider multimodal transportation and focus on route suggestions for drivers only.

In~\cite{Eiter2014} a system of semantically-enriched personalized public-transport routing is 
presented. Individual preferences are learned from previous travel behavior using a 
multinomial mixed logit, where individual-level parameters (see, e.g.,~\cite{train03}) are exploited 
to generate personalized public-transport route choice models.
However, neither initialization nor belief updating mechanisms are described there. 

The integration of real-time traffic information with customized route recommendations is considered, e.g., in~\cite{tourism14}, which introduces a personalized-route recommendation system for tourists. Tourists that are planning to visit $n$ points of interest are recommended paths that can guide them to complete their trip. Recommendations are generated  by trading off personal interests and visiting preferences of the tourists against current traffic conditions, in order to prevent  congestion and queues. The approach in~\cite{parks12} collects data about the visiting behavior of tourists in museums and park themes. Based on this information, tourists are clustered into different groups. New tourists are assigned to a cluster and recommended paths based on the preferences of users in the same cluster and on real-time queuing conditions.
However, these approaches relate to scheduling of intermediate goals in a rather restricted setting of a theme 
park and hence are not directly applicable in our setting. 
The main conclusion of the survey~\cite{Gavalas2014} particularly related to tourism applications is that 
a main open questions consists of the representation, estimation and 
assessment of individual preferences. Our work addresses this open question. 

Recently, crowd-based route-recommendation systems have also been proposed~\cite{crowdsourcing14,crowdplanner14}. They resorts to crowd knowledge to improve route-recommendations quality. In particular, they ask human workers for evaluating the alternative routes suggested by a navigation system for a given point-to-point trip. The feedback from the crowd is used to help the traveler in choosing a route from the set of alternatives. These approaches share with FAVOUR the assumption that traveling preferences may be transferred among users and that traveling suggestions for a single users can be improved from the feedback of other travelers. 
However, these approaches focus on route choice and neglect mode choice. Also they do not 
include situation awareness. Moreover, they do not provide any strategy on the 
intialization of the personal profile which is the focus of this paper. 

The work closest to FAVOUR is 
the strategy described in the recent paper~\cite{Nuzzolo2015a} using the methods of~\cite{personalizedITS_2013}. It tackles the same problem addressed by FAVOUR, the design of 
personalized multimodal route-recommendations systems, and shares with FAVOUR the incremental Bayesian 
approach for learning the traveler preferences. The experiments reported in~\cite{personalizedITS_2013} show that the Bayesian approach can learn the preferences of the users, 
in the case of 
simulated choice scenarios with three different transportation alternatives 
(public transport, car, public transport and car). 
The goal of our work is however \emph{different}: we show that preferential knowledge can be 
\emph{transferred} among users, in order  to improve the learning of their utility models, in 
particular at the initialization stage. 
That is, \textit{transfer learning}, which has been shown to provide added benefits when 
solving a series of related problems~\cite{transferLearning10}, can improve personalized-route recommendations systems.

\section{Problem setting}
\label{sec:ps}

A personalized, situation-aware route-recommendation system learns the travel preferences of the users and exploits them to drive the search of the underlying routing system for a satisfacing route. This section describes both the requirements imposed to the learning system by the interaction with the travelers and the 
structure of the routing model underlying the FAVOUR algorithm. 

\subsection{Efficient interaction with the travelers}

A personalized situation-aware routing service faces several challenges in terms of 
interaction with the users, stemming 
from the bounded rationality of the humans when making decisions~\cite{boundedrationality_78}:
\\
1) \emph{robustness to noisy feedback}~\cite{BCEMOrob,PEprinciples10}. The typical human decision-making process is characterized by imprecision, contradictions, and changes of judgement over time;
\\
2) \emph{low cognitive load of the user}~\cite{PEprinciples10,BCEMO,compQueries09,comp_and_fatigue_00}. 
While the users can provide qualitative evaluations, such as ``I prefer route A to route B'', they usually cannot quantify the extent to which they prefer route A over route B. Therefore, quantitative feedback from  the user is not suitable in the context of a traveling information system;
\\
3) \emph{real-time constraints}~\cite{PEprinciples10}. 
A traveler usually expects an ``immediate'' recommendation from the system after posing a query. Therefore, the suggestion of new alternative routes cannot take more than few seconds,  
while the refinement of the user preference model can be done in an offline process during inactivities of the user;
\\
4) \emph{user perception of the interactive process}. 
If the quality of the recommended routes does not increase when the user provides additional information, she is likely to get annoyed, perceiving the system as useless, and to stop using it.

\subsection{Underlying routing model}

FAVOUR assumes that routing decisions are taken according to a random utility maximization 
based on systematic utilities that use additive 
as well as non-additive path costs in the sense of~\cite[see sections 2.3.3. and 4.3.3]{Cascetta}.
Thus FAVOUR considers routing on a multimodal transport network, which can be represented by a 
joint graph $G=(V,E)$.
Edges $e \in E $ and nodes $v \in V$ in the graph are described by a set of attributes, and each edge (node) is associated with a set of attribute values
such as the edge length and 
travel time. While the graph is considered static the attributes may depend on the current situation and the context.
A route $\Br$ is defined as an ordered sequence of adjacent edges $(e_1, e_2, \dots , e_l)$ from the starting to the destination node, with $l$ denoting the 
number of edges of the route. The 
systematic route cost $U(\Br)$ of a candidate route $\Br$ can therefore be defined as follows: 
\begin{equation}
	\label{eq:route_utility}
	U(\Br)= U_{mw}(\Br)+\sum_{i=1}^{l} C_i
\end{equation}
where $C_i$ denotes the cost function for edge $i$, which depends on the 
attributes $z_{i,j}$ of the edge $e_i$ in the following form:
\begin{equation} 
	C_i = \sum_{j=1}^m w_j c_j(z_{i,j})
\end{equation}
The edge costs $c_i$ thus are a weighted linear combination of $m$ \emph{basis} cost functions 
$c_j: \mathbb{R} \mapsto \mathbb{R}^+ \cup \{0\}$.  
This formulation enables a non-linear dependency between the utility of an edge and the edge 
attributes. 
One example of interest in this respect could be the incorporation of nonlinear forms of delay 
terms that introduce an additional penalty for edges with long delays to account for the 
corresponding frustration. 
The function $U_{mw}(\Br)$ enables the incorporation of nonadditive route costs related to the modes used in the route and to the current weather conditions. It acts as an alternative-specific constant. 

Let the vector $\w = [w_j]_{j=1,...,m}$ denote the parameters (or 
weights) of the linear combination.
Then   
Eq.~(\ref{eq:route_utility})  can be reformulated as follows:
\begin{equation}
 	\label{eq:route_utility_dp}
 	\begin{aligned}
	    U(\Br) & = U_{mw}(\Br) + \sum_{i=1}^{l} \sum_{j=1}^{m} w_j  c_j(z_{i,j}) \\
 		 	  & = U_{mw}(\Br) +\w \cdot \u(\Br)    
  	\end{aligned}
\end{equation}
where the components of the vector $\u(\Br)$ are defined as: 
$$
u_j(\Br) =  \sum_{i=1}^{l} c_j(z_{i,j}).
$$
In the following the vector $\w$  is called {\em profile} and is allowed to be specific for each user. 

Let us note that the route choice corresponding to the above formulation of route costs is quite flexible. It allows 
for situation dependence in 
terms of edge costs corresponding to real time expected travel times by letting $z_{i,j}=z_{i,j}(t)$ depend on the time  accounting for disruptions
in the transport system such as, e.g., unexpected traffic jams due to construction work or accidents in the public transport system.
Also $U_{mw}(\Br)$ can account for a great number of different personal preferences, including favorite modes and mode/weather conditions.

\section{Favourite route-recommendation algorithm}
\label{sec:FAVOUR}

As detailed in the introduction, FAVOUR uses three stages for the personalization of the weight vector $\w$ 
(in the following called {\em user profile}) 
encoding 
the preferences of a user. In the first stage, general sociodemographic questions are asked in order to classify the
new user in a small number of user classes by separating, e.g., students from old-aged people or workers from unemployed. 
For each of these classes a {\em mass preferences model} is estimated representing the average tastes in this class. 
Additionally in the first stage mobility restrictions are queried relating to the mobility options of the user as well as to significant places such as home and work location. 

Based on this rough first guess stated preference questions in the form of realistic routing situations are asked in 
the second stage using the knowledge from the first stage, as is usual for personalized route recommendation systems~\cite{Nuzzolo2015a,personalizedITS_2013}. 
In this respect FAVOUR adopts
simple comparison queries 
of the kind ``which of these two candidate routes do you prefer?'' 
(see Sec.~VI below).
Comparison queries are well-known in the preference learning literature to be much more affordable  for the users than quantitative judgements~\cite{compQueries09}.
Based on the answers the weight vector $\w$ is adapted to personal preferences. In stage three this personalization process is continued based on real path choices observed in reality. 

Personalization is achieved by FAVOUR through the learning mechanism presented in the following.  
Sec.~\ref{sec:cold-start} then describes a strategy to transfer knowledge among related preference learning tasks and integrates it in the learning algorithm.

\subsection{Learning a utility model from user preferences}
\label{subsec:bl}

FAVOUR adopts a Bayesian strategy in order to infer the personal preferences encoded in vector $\w$. 
Given the prior belief  $\mathrm{p}(\w)$, i.e., the initial belief without having seen any user preference, and the likelihood model $\mathrm{p}(T | \w)$ linking the training preference information $T$ to the random variable $\w$,  the posterior belief $\mathrm{p}(\w | T)$ is inferred. 

\vspace{0.2cm}\noindent
\textbf{Prior distribution. } 
In the following we argue that initialization with 
a special prior
specific to each user class, 
called {\em mass preference prior} (MPP), is beneficial. 
Details on the calculation of the MPP are given in Sec.~\ref{sec:cold-start}. 

\vspace{0.2cm}\noindent
\textbf{Likelihood model. } The user expresses her preferences  by pairwise comparisons of candidate routes.  
Therefore the training set $T$ consists of pairs $(\Br_t, \q_t), t=1,...,|T|$, with route $\Br_t$ being preferred to 
route $\q_t$ (denoted as  $\Br_t  \succ \q_t$) and the probability thereof for given $\w$ being modeled as: 
\begin{equation}
	\label{eq:link_function}
 	\mathrm{Pr} (\Br_t  \succ \q_t | \w) =  \frac{1}{ 1 + e^{U(\q_t)  -  U(\Br_t)} } %
\end{equation}
That is the logit model is chosen for the binary comparisons. Other choices such as the probit model are
possible and straightforward to include, however, as the performance of probit and logit models for binary choice 
is comparable~\cite{pairwiseComp11}, we here opt for the computationally simpler logit model. 
Adding independence over choice situations given the personal preferences leads to the likelihood:
\begin{equation}
	\label{eq:BT_lik}
 	\mathrm{p}(T | \w) = \prod_{t=1}^{|T|} \mathrm{Pr} (\Br_t  \succ \q_t | \w) = \prod_{t=1}^{|T|} \frac{1}{ 1 + e^{U(\q_t)  -  U(\Br_t)} }
\end{equation}

\vspace{0.2cm}\noindent
\textbf{Inference technique. }
With the selected likelihood, exact Bayesian inference is not analytically tractable and approximations of the posterior distribution $\mathrm{p}(\w | T)$ are thus necessary. 
To generate real-time recommendations, deterministic approximate inference is adopted,  as it is computationally cheaper than stochastic approaches, which approximate the posterior by repeatedly drawing independent samples from it. 

A computationally-affordable deterministic approximation is computed by the Laplace inference technique
that generates a Gaussian approximation $\mathcal{N}(\w | \tilde{\boldsymbol{w}}, \tilde{\boldsymbol\Sigma})$ of the posterior.
The vector $\tilde{\boldsymbol{w}}$ is the mode of the un-normalized posterior $ \mathrm{p}(T | \w)  \times \mathrm{p}(\w)$ , providing the ``most probable'' a-posteriori weight vector $\tilde{\boldsymbol{w}}$.
The covariance matrix $\boldsymbol\Sigma$ is equal to $(-\boldsymbol{H})^{-1}$, with $\boldsymbol{H}$ being the Hessian matrix 
of the un-normalized log-posterior computed at $\tilde{\boldsymbol{w}}$. 
Therefore, the Laplace method provides a quadratic approximation to the log-posterior around its
mode.

In general, the Laplace approximation may be unrepresentative of the overall posterior
mass, especially for multimodal posteriors. More sophisticated approximate inference techniques exist~\cite{Bishop06} to handle multimodal posteriors. 
However, these techniques are computationally more expensive than the Laplace method, and therefore inappropriate for our real-time task.
Most important,  our choice of the Gaussian prior and of the Bradley-Terry likelihood  guarantees the unimodality of the posterior to be approximated. 
In addition to unimodality, the log-concavity property of the posterior also guarantees that the tails of the posterior are no heavier than $e^{-|w|}$, 
providing arguments for the Gaussian approximation. %

In general, the domain of the weights 
$w_j$ is $\mathbb{R}$, to encode both what the user likes (positive values) and what she dislikes (negative values). However, assigning positive weights to some specific decisional features 
(like, e.g., travel time) is not reasonable. 
While traditional approaches for constrained weights rely on bounded-domain priors (e.g., the exponential distribution for positively-constrained weights~\cite{L1_12}), in this work, for computational reasons, we constrain the posterior mean to lie in a box-bounded set by restricting the maximization of the log-posterior 
$\log \mathrm{p}(T | \w)  + \log \mathrm{p}(\w)+c(T)$. 
\subsection{Incremental learning strategy}
\label{subsec:il}

In FAVOUR the posterior is computed incrementally, see the pseudo-code presented in 
Fig.~\ref{alg:FAVOUR}. Here $T^{i}$ denotes the training set obtained after asking $i$ questions. 
The prior distribution $\mathrm{p} (\w | T ^{i-1})$ over $\w$ after $i-1$ questions is updated 
once the $i$-the question is answered by applying  the Bayes rule 
\begin{equation}
\label{eq:incremental_belief_update}
\mathrm{p}(\w | T ^{i}) = \frac{ \mathrm{p}(T ^i | \w)  \times \mathrm{p}(\w | T ^{i-1})  } { \mathrm{p}(T ^i) } 
\end{equation}
to result in the posterior  $\mathrm{p}(\w | T ^{i})$. 
The posterior is computed by applying the inference procedure described above. The iterative learning in stage two is 
stopped if a termination criterion is met. 
Such a criterion may include the number of questions asked, the accuracy achieved or the time used for the stage.

\begin{table*}[tb]
\begin{tabular}{p{0.5\textwidth}p{0.5\textwidth}} 
\centering
\begin{small}
	\begin{pseudo}
 	\textbf{procedure} Incremental Learning \xl
	\xn  \textbf{input:}   transportation system graph \xl
	\xn  \textbf{output:} approximation $\tilde{\w}$ of the user preference model  $\w_*$ \xl
	\xn \textbf{Let} $T=\emptyset$  training set of the user preference information 
	\xl
\xs{0.5cm}
    \xn {\bf do} \xl
    \xn  \xn  \xa \textit{/* Refinement phase */} \xl
    \xn  \xn   \xb Based on set $T$, update the current belief $\mathrm{p} (\w |T)$  \xl
		\xn  \xn   \xb  \textit{/*  Bayesian inference */} \xl
    \xn  \xn  \xc  \textbf{Let} $ \tilde{\w}=\mathrm{argmax}_\w \mathrm{p} (\w |T) $ \xl 
		\xn  \xn  \xc /* Mode of the posterior distribution */  \xl 
		\xn  \xn  \xc and calculate $\tilde{\Sigma}$ /* variance of posterior */ \xl
    \xn  \xn  \xa \textit{/* Preference elicitation phase */} \label{line:pe} \xl
    \xn  \xn   \xb Ask query $t$  to the user \xl
    \xn  \xn   \xb Get user response to query $t$ \xl
    \xn  \xn   \xc Include the training example from the user response \xl %
 \xn {\bf until} (termination\_criterion) \xl
    \xn \textbf{return} approximation $\tilde{\w},\tilde{\Sigma}$ of the user preference model  $\w_*$
  \end{pseudo}
\end{small} 
  \captionof{figure}{Pseudo-code of the iterative learning approach for an approximation $\tilde{\w}$ of the unknown preference model $\w_*$.}
  \label{alg:FAVOUR}
&
\centering
\begin{small}
	\begin{pseudo}
 	\textbf{procedure} MPP refinement\xl
	\xn  \textbf{input:}  training sets $T_k, k = 1 \dots,  K$ for the $k$ users \xl
	\xn  \textbf{output:} mass-preference prior MPP \xl

	\xs{0.5cm}
    	\xn  \textit{/* Initialization phase */}\xl
    	\xn  \textbf{Set} MPP to  $\mathcal{N}(\boldsymbol{0}, \boldsymbol{I})$, with $ \boldsymbol{I}$ being the identity matrix   \xl	
    	\xn \textit{/* Refinement phase */}\xl	
    	\xn {\bf do} \xl
    	\xn  \xn    Based on MPP, %
			update the user utility models \xl
			\xn  \xn   $\mathcal{N}(\boldsymbol\mu_k, 	 \boldsymbol\Sigma_k), k = 1 \dots,  K$ \textit{/*  Bayesian inference */} \label{line:bi} \xl
    \xn  \xn    Compute $\bar{\boldsymbol\mu}$ and $\bar{\boldsymbol\Sigma}$ according to Eq.~(\ref{eq:MPP}) \xl
    \xn  \xn   Set MPP to  $\mathcal{N}(\bar{\boldsymbol\mu}, \bar{\boldsymbol\Sigma})$   \xl
 \xn {\bf until} (termination\_criterion) \xl
    \xn \textbf{return} mass-preferences prior MPP
  \end{pseudo}
\end{small} 
  \captionof{figure}{Pseudo-code for iteratively learning the mass-preferences prior. Each user's utility model (line~\ref{line:bi}) is updated by applying Bayesian inference.} 
  \label{alg:MPP_ref}
\end{tabular} 
\end{table*}

\section{Transferring knowledge among utility models}
\label{sec:cold-start}

The task of efficiently learning the preferences of new users is henceforth referred to as the \emph{cold-start problem}.
To decrease the number of queries asked to new users, FAVOUR exploits the preferences 
elicited from \emph{past} users. Transferring knowledge among related learning tasks is known as transfer learning~\cite{transferLearning10} in the machine-learning community. In this paper transfer learning methods are used in order 
to infer priors for user classes identified using the stage one questions. For each user class 
a {\em mass preference prior} (MPP) is learned from experiences of past users in this class. 
The main idea in this
respect is that the MPP should encode the typical behavior of a given class of users 
such that if new answers from user in the class are incorporated into the
MPP, this should not be changed.

In particular the computationally affordable method introduced in~\cite{PL_multiple_DMs_12} is adopted.
It computes a  \emph{Gaussian MPP} $\mathcal{N}(\bar{\boldsymbol\mu}, \bar{\boldsymbol\Sigma})$ 
by  averaging the posterior Gaussian models $\mathcal{N}(\boldsymbol\mu_k, \boldsymbol\Sigma_k), k = 1 \dots,  K$ of $K$ users as follows:
\begin{equation}
 	\label{eq:MPP}
 	\begin{aligned}
		\bar{\boldsymbol\mu}   & =           \frac{1}{K} \sum_{k =1}^{K} \boldsymbol\mu_k \\
 		\bar{\boldsymbol\Sigma} & =      \frac{1}{K} \sum_{k =1}^{K} (( \boldsymbol\mu_k -  \bar{\boldsymbol\mu}) ( \boldsymbol\mu_k - \bar{\boldsymbol\mu} ) ^T +  \boldsymbol\Sigma_k)
  	\end{aligned}
\end{equation}

In~\cite{PL_multiple_DMs_12}, an iterative procedure is introduced to refine the learned MPP, which is described in the form of a pseudo-code in Fig.~\ref{alg:MPP_ref}.
In particular, the MPP learned at the iteration j-1 (by applying Eq.~(\ref{eq:MPP})) 
is used at the j-\emph{th} iteration as prior for the Bayesian inference refining the utility model of each user.  
The refined utility models are then exploited to re-estimate the MPP.
The incremental procedure for MPP learning %
is iterated until convergence. 
In our work, the difference among MPPs is measured by the Kullback-Leibler (KL) divergence for two probability distributions. Convergence is numerically detected if the KL-divergence of consecutive iterations differ by less than a threshold. 
As the algorithm is guaranteed to converge~\cite{PL_multiple_DMs_12}, the process will stop. 
Also note that the calculation of the MPP can be done offline and hence time restrictions are not severe in this respect. 

We have also considered more sophisticated approaches %
for generating the MPP. In particular, the user utility models have been clusterized into a set of clusters, each one being represented by a Gaussian distribution. The mixture of Gaussian distributions defining the clusters has been used as MPP. 
However, this
approach did not yield any relevant performance improvement
during the experimental evaluation. 

\section{Experimental results}
\label{sec:exp}
The proposed algorithm for stage two described above has been tested in a real-world case study involving 40 participants 
including people of different gender, age, profession (students, workers or retired persons) and socioeconomic status.
The participants are not assigned to different classes but only one MPP is estimated for all
participants despite their heterogeneity. 
First, the performance of the FAVOUR algorithm adopting the mass-preferences prior is compared with the results observed when the mass-preferences prior is \emph{not} used.  
For this purpose, the travel choices made by a set of 
\emph{real} decision makers in \emph{realistic} traveling situations have been collected. The resulting dataset is described in Sec.~\ref{subsec:dataset} and~\ref{subsec:routeFeats}.   
Details of the FAVOUR implementation are provided in Sec.~\ref{subsec:FAVOURimpl}, while an alternative non-Bayesian transfer-learning approach is defined in Sec.~\ref{subsec:alternative}.
The benefits of adopting the mass-preferences prior to initialize the incremental learning algorithm of FAVOUR are shown in Sec.~\ref{subsec:tl_benefit}.
Finally, in Sec.~\ref{subsec:FAVOUR_vs_ml_pl} and~\ref{subsec:FAVOUR_vs_sta}, the efficiency of the transfer learning technique used by FAVOUR is evaluated, respectively, by a comparison with:
\begin{itemize}
	\item the alternative non-Bayesian transfer-learning approach described in Sec.~\ref{subsec:alternative};
	\item the state-of-the-art of route-recommendation system~\cite{personalizedITS_2013} discussed in the related work section.
\end{itemize}
The main questions to be answered empirically relate to the achievable accuracy and the number of questions that need to be asked in order to attain reasonably accurate recommendations.

\subsection{Realistic travel-choices dataset}
\label{subsec:dataset}

The travel choices are collected in the form of a stated preference (SP) survey 
in line with common practice in the literature~\cite{personalizedITS_2013,Nuzzolo2015a}.
However, like in the proposed three stage system, the SP questions are personalized for each respondent to get informed decisions. While there are still some concerns regarding the validity of SP surveys when it comes to realistic parameter values, the personalization should improve the results. 
In our experiment, first for every interviewee five route choice situations (RCS) are defined.  Each RCS involves three alternative routes to be compared by the interviewee. Each of the five situations is replicated with a different setting of weather conditions, for a total of ten RCS to be tackled by the interviewee.
Ternary rather than binary choices are asked in the SP surveys, in order to reduce the total number of questions, while still holding the number of alternatives to be evaluated small. The ranking of the three alternatives from one  interviewee generates three binary choices, used in the second stage of FAVOUR for modeling.

The route alternatives provided to the participants in each RCS are personalized according to their available transportation modes, in order to present each participant with valid alternatives. 
Furthermore real-world travel-times predictions 
thus including average congestion delays 
are used that are specific to the day and the time 
given in the route choice questions. 
The following individual and public transportation modes are considered:
\begin{itemize}
	\item car driving;
	\item car driving (car sharing);
	\item cycling;
	\item cycling (bike sharing);
	\item walking;
	\item public transport (bus, underground or tram).
\end{itemize}
Routes within the city of Vienna are generated for combinations of the above transportation modes. Public transportation routes are obtained by querying the electronic router of the Viennese local public transport provider. For the individual modes, the multimodal router Ariadne~\cite{prandtstetter2013way} 
using real-world real-time 
travel-times is exploited.

To guarantee a founded and informed ranking by the interviewee, the alternative routes are visualized in a map and either the destination or the origin in up to four of the ten choice situations is the actual interviewee's home address. The remaining destinations and origins are selected within well-known areas of Vienna.
The choice situations generated are described by using the set of route features discussed in 
the next section. An example of the customized and detailed routes presented to the interviewees can be seen in Fig.~\ref{fig:questionnaire}.

\begin{table*}
\begin{tabular}{m{6.5cm}m{9cm}}
	\begin{tabular}{|l|p{4.5cm}|}
	\hline
\multirow{5}{1.2cm}{\textit{Distance \newline features}} 
		& dist. covered by driving the car \\
		& dist. covered by driving the car over main roads \\
		& dist. covered by walking \\
		& dist. covered by cycling \\
		& dist. covered by cycling within cycling infrastructure \\
	\hline	
    \multirow{10}{1.2cm}{\textit{Time \newline features}} 
		& time spent by walking \\
		& time spent by cycling \\
		& time spent by driving \\
		& access walking time within PT station \\
		& egress walking time to get out of PT station \\
		& time spent by waiting for the PT mean\\
		& time spent in bus \\
		& time spent in tram \\
		& time spent in metro \\
		& time spent to switch among non-PT means / to park the car \\
	\hline
	\multirow{3}{1.2cm}{\textit{Cost \newline features}} 
		& driving cost \\
		& parking cost \\
		& PT-ride cost \\
	\hline
    \multirow{5}{1.2cm}{ \textit{Miscellaneous \newline features}} 
		& number of transportation-means changes\\
		& uphill altitude gap filled by cycling \\
		& number of bus stops \\
		& number of tram stops \\
		& number of metro stops \\
	\hline
	\end{tabular}
	\captionof{table}{Decisional features corresponding to a route based on attributes of the edges. 
The numerical values of the features are collected in the features vector $\u(\Br)$. The acronym ``PT'' stands for public transportation.}
\label{tab:dec_feats}
&
	\centering
	\includegraphics[width=.6\textwidth]{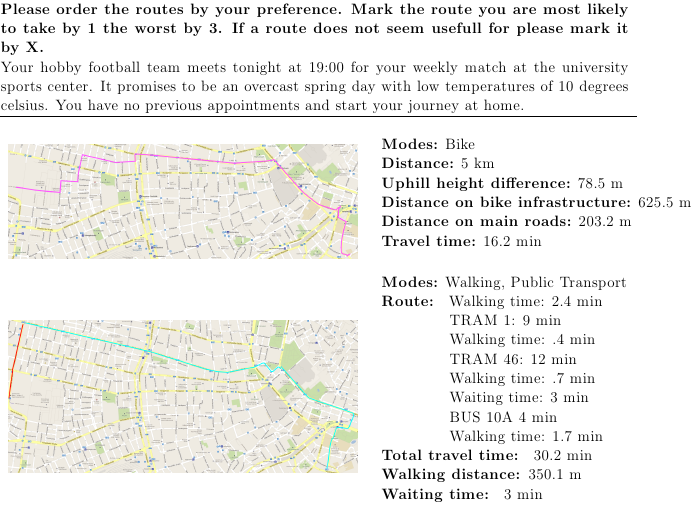}
	\captionof{figure}{Two of the three routes in one of the ten scenarios presented to the participants of the survey (the third route is not shown due to space restrictions). The  routes in this scenario start at the actual home address of the participant and lead to the well-known University sports center. 
	The survey was provided in German (translation by authors only for paper).}
	\label{fig:questionnaire}
\end{tabular}
\end{table*}

\subsection{Route features}
\label{subsec:routeFeats}

The constant $U_{mw}(\Br)$ specific to the routes and the situational context provided to the interviewees for evaluation is described by a set of binary decisional features generated as follows.  
Each transportation mode is associated with an indicator variable, set to 
one if the transportation mode is used to cover (a part of) the route. As this study includes 
multimodal routes, different indicator variables may be simultaneously set to one. 

The environmental conditions are classified into six different states, each one 
encoded by a dummy indicator variable. The states describe the six possible 
interaction terms of the precipitation-chance binary-forecast (high chance of rainfall or snowfall w.r.t. no precipitation chance) with three different temperature intervals 
(low, medium and high temperature).

Each of the six environmental-conditions variables are combined with the six different transportation modes, 
thus generating 36 binary indicator variables $m_j(\Br,\s)$. The indicator variables define the transportation modes used with certain environmental conditions represented in the vector $\s$. 
For example, one variable indicates whether the route includes using the car with high chance of precipitation and low temperature.
Combining these features we obtain
$$
U_{mw}(\Br,\s) = \sum_{j=1}^{36} w_j m_j(\Br,\s).
$$

In addition to the 36 possible combinations of transportation modes with environmental conditions,
we consider also the time to cover the route, the route length, the travel cost and the number of transportation-means changes, encoded in $u_j(\Br), j=37,...,59$ (see Tab.~\ref{tab:dec_feats}). 
In detail, we measure the time spent 
(specific to the given day and time of the day) 
and the distance covered with each transportation mean. 

Specific features of transportation modes are also considered: for bike, this contains the percentage of the distance covered within cycling lanes and the filled uphill altitude gap;
for public transportation, the egress/access/switching time, the ride total cost and the number of stops for each transportation mean; for car, the distance covered over main roads, the driving and parking costs. 
The fine granularity used in selecting the route features 
allows some flexibility in encoding the decisional process of a generic user.

\subsection{FAVOUR implementation}
\label{subsec:FAVOURimpl}

The incremental learning component of FAVOUR is implemented in Matlab R2008a.  
The box-bounded optimization task identifying the posterior mode $\tilde{\boldsymbol{w}}$ 
and variance $\tilde{\Sigma}$ is solved by a  trust-region optimization algorithm, based on  local quadratic approximations of the non-linear objective function. As matter of fact, the Hessian matrix is also needed to generate the covariance matrix $\boldsymbol\Sigma$ for the Laplace inference technique. The analytic formulation of the log-posterior derivatives has been provided to the optimization algorithm. In fact, numerical differentiation techniques may in general be unstable and computationally expensive for \textit{real-time} route recommendations. 
Even though the objective function of the optimization task is unimodal and thus a single run of the local-search algorithm can identify the posterior mode, robustness w.r.t. numerical inaccuracies is increased by performing five runs of the algorithm. The starting point of each local-search run is selected uniformly at random, with the seed of the random number generator set to one.

Let $\Br$ and $\q$ be a pair of routes. The predictive distribution for $\Br \succ \q$, given the routes feature vectors  $\u(\Br)$, $\u(\q)$, is obtained by marginalizing w.r.t. the posterior distribution $\mathrm{p}(\w | T)$:
\begin{eqnarray}
	\label{eq:app_prediction}
    & \mathrm{Pr} (\Br  \succ \q | \u(\Br), \u(\q), T) \approx 	\nonumber	\\ 
    & \int s(U(\Br) - U(\q)) \mathcal{N}(\w | \tilde{\boldsymbol{w}},\tilde{\boldsymbol\Sigma}) \mathrm{d}\w 
\end{eqnarray}
where $s(x)$ is the logistic sigmoid $1/(1 + e^{-x})$ of the Bradley-Terry likelihood model and $\mathcal{N}(\w | \tilde{\boldsymbol{w}},\tilde{\boldsymbol\Sigma})$ is the Gaussian approximation of the posterior generated by the Laplace inference technique. 
A well-known approximation of Eq.~(\ref{eq:app_prediction}) consists of replacing the logistic with the probit sigmoid~\cite{Bishop_book06}, as convolving a probit sigmoid with a Gaussian results in another probit sigmoid. By setting $\d = \u(\Br) - \u(\q)$, the final result is the following computationally-affordable formulation~\cite[cf. pp. 218--220, Chap. 4]{Bishop_book06}:
\begin{equation}
 	\label{eq:final_app_prediction}
	\mathrm{Pr} (\Br  \succ \q |  \u(\Br), \u(\q), T) \approx s(\lambda \tilde{\boldsymbol{w}}^T \d )
\end{equation}
where:
$$\lambda \ = (1 + \pi \d^T  \tilde{\boldsymbol\Sigma} \d / 8)^{-1/2}$$
Equation~(\ref{eq:final_app_prediction}) is used during the testing phase (see next sections) to predict whether the user prefers route $\Br$ to route $\q$. The probability value provides an estimate for the predictive uncertainty of FAVOUR.

\subsection{Alternative transfer-learning algorithm}
\label{subsec:alternative}

For a learning algorithm alternative to the one of FAVOUR, we follow the methodology of~\cite{Eiter2014}, applying the idea of individual level parameters %
in mixed logit 
models~\cite{train03}. The utility of a route $\Br$ of user $k$ with $l$ edges is the same as the one 
in Eq.~(\ref{eq:route_utility_dp}) using the same notation:

\begin{equation*}
	    U(\Br)  =  U_{mw}(\Br) +\w \cdot \u(\Br)    
\end{equation*}
To allow for individual level parameters we assume a mixed logit formulation, where $w_j\sim  \mathcal{N}(\mu_j, \sigma_j)$ (independent over components). 
In a first step the parameters $\mu_j$ and $\sigma_j$ are estimated using a simulated maximum likelihood approach. 
With the currently used sample sizes the unrestricted mixed logit model contains too many parameters and 
model selection methods are used in order to obtain a better 
model for the average preferences. 
In this respect a fast forward-backward variable selection procedure based on the Aikake Information Criterion (AIC) is used. In a first step a forward variable selection is applied, adding one variable at a time. Afterwards the variable selection is fine-tuned by a backward-forward procedure, which adds/deletes single variables from the model as long as the AIC keeps improving. 
This procedure results in a reduced model with a smaller number of variables. 

To ensure that variables  relevant to the decision maker are not missed in the personalized model, the variables excluded during variable selection are added into the personalization procedure with zero mean and variance value comparable to the variance value of the included parameters. The resulting mixed parameter model serves as an input to a second step, where individual level parameters are estimated by ensuring that parameters relevant for the decision maker are assigned  values different from zero, while irrelevant parameters are not. 
The estimation procedure of the individual level parameter values works as follows~\cite[cf. p. 300, Chap. 11]{train03}:
\begin{enumerate}
	\item for the estimation of the individual level parameters we draw a sample
	$\w^{(b)}$ of size $B$ from  
	$\mathcal{N}(\hat \mu, \hat \Sigma)$ which is the estimated distribution of the
	weights $\w$ in the mixed parameter utility model estimated above;
	\item the simulated subpopulation mean is calculated as
	\begin{equation*}
		    \check{ \w} = \sum_{b=1}^B g_b \w^{(b)}  %
				= \frac{\sum_{b=1}^B \w^{(b)} \prod_{k \in
	K}P(y_{\Br_k}|U({\Br_k}),\w^{(b)})}{\sum_{a=1}^B  \prod_{k \in
	K}P(y_{\Br_k}|U({\Br_k}),\w^{(a)})}
	\end{equation*}
	where $P(y_{\Br_k}|U({\Br_k}),\w^{(b)})$ is the probability that the individual
	chooses route $\Br_k$ for the $k$-th question of the survey. Furthermore: 
	\begin{equation*}
	P(y_{\Br_k}=1|U({\Br_k}),\w^{(b)})=\frac{exp(U(\Br_k))}{\sum_{\Br_m} exp(U(\Br_m))}
	\end{equation*}
	where the sum in the denominator is over all routes involved in the decision. 
\end{enumerate}
The resulting simulated weights $\check{ \w}$ define the user profile which can be
exploited to obtain the most probable choice.

\subsection{Evaluating the benefit of transfer learning}
\label{subsec:tl_benefit}

The efficiency of FAVOUR is demonstrated by a \emph{leave-one-user-out cross-validation} (LOUO-CV) procedure. In each run of the cross-validation, a single user is selected as test user and a mass-preferences prior is learned over the remaining $39$ (training) users. 
The 30 preference examples of the test user are split into five test examples and $25$ candidate training examples. 
Five training sets of size $2,4,6,8,10,12,15$ are generated by sampling uniformly at random the $25$ candidate training examples. 
A utility model for the test user is learned from each training set and its predictive accuracy is evaluated over the test set. To get stable results, the 30 preference examples of the test user are re-partioned five times into the test set and the candidate training examples set. The whole training procedure considering an increasing number of training examples is repeated for each partition. 
The learning of an utility model for a \emph{specific} test user over a \emph{specific} training set instance and its evaluation by using a \emph{specific} test instance defines a \emph{test session} for the FAVOUR algorithm.

\begin{figure}%
	\centering
	\includegraphics[width=.5\textwidth]{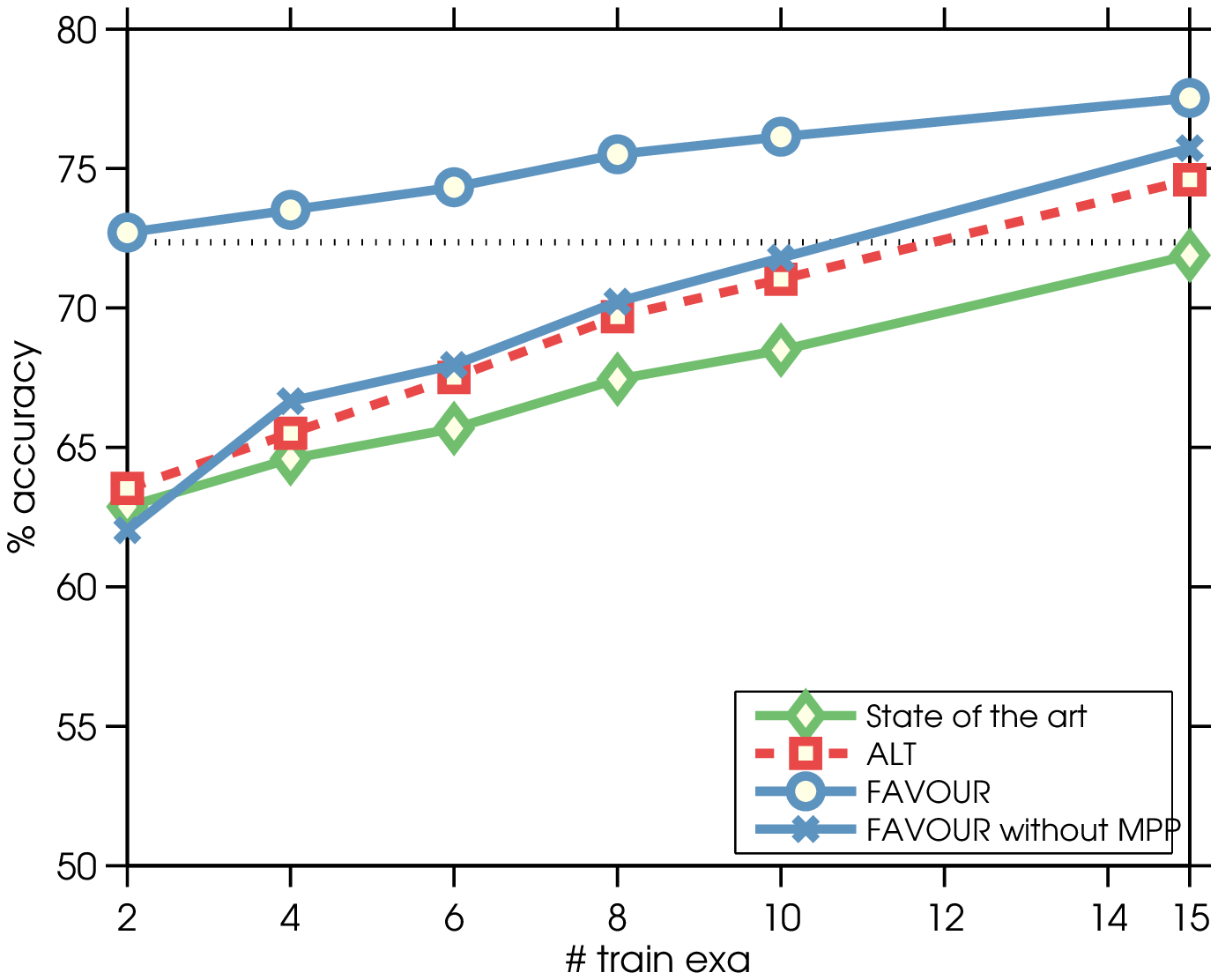}
	\caption{Learning curves for the different algorithms: the FAVOUR algorithm with MPP (solid curve, `o') or without (solid curve, `x'), the mixed logit model (ALT, dashed curve), and the simulation of the  algorithm~\cite{personalizedITS_2013} (solid curve, diamond markers). 
\textit{y}-axis: the mean predictive accuracy (percentage values). \textit{x}-axis: the number of training examples.
Dotted line: 72.3\% (no personalization, MPP).} 
	\label{fig:MPPperformance}
\end{figure}

The utility model of the test user is learned by executing the Bayesian learning strategy of FAVOUR (pseudo-code in Fig.~\ref{alg:FAVOUR}) initialized with:
\begin{itemize}
\item the uninformative Gaussian prior $\mathcal{N}(\boldsymbol{0}, \boldsymbol{I})$ (solid curve with cross markers in Fig.~\ref{fig:MPPperformance});
\item the mass-preferences prior computed over the training users (solid curve with circular markers).
\end{itemize}
The \emph{x}-axis in Fig.~\ref{fig:MPPperformance} reports the size of the training set, while the \emph{y}-axis shows the accuracy. Each point in the graph represents the mean percentage accuracy averaged over all test sessions performed for all possible test users.
For each test session, the accuracy is computed as the percentage of correct predictions over the five test pairwise comparisons. The \naive\ baseline for the predictive performance is thus the 50\% accuracy obtained by random guessing, which is the value of the lower bound for the \emph{y}-axis.

Confidence intervals for the percentage accuracy mean are not shown in the graph to avoid cluttering. The dispersion of the observed accuracy values is however discussed in the supplemental material (Sec.~\ref{sec:supp}). 

The graph clearly depicts the performance improvement gained by adopting the MPP. In detail:
\begin{itemize}
\item the benefit of adopting the MPP is 
demonstrated by the superior performance observed for all training set sizes. 
Only for $15$ training examples, when on average enough preference information is elicited from the new user, the MPP is only slightly better than the alternatives.
The accuracy improvements due to the adoption of the mass preferences prior are statistically significant, according to  
a one-sided Kolmogorov-Smirnov test.
 Six statistical tests are performed, one for each of the six values of the training set size $s$.  A Bonferroni-corrected significance-level $8e-3$ is used, obtained by dividing the standard significance level $0.05$ by the total number of tests.  For $s=2,4,6, \dots,10$,  the order of magnitude of the p-values is lower than or equal to $5e-07$, providing strong evidence for a statistically different performance;
\item the fewer training examples are used, the more pronounced is the benefit of adopting the MPP.  With two and four training examples only, a 10.6\% and 6.8\% accuracy difference is observed, respectively. When the training set size is 10, the accuracy difference decreases 
to 4.3\%,  eventually becoming statistically null when 15 examples are used;
\item if the MPP is not used, a faster accuracy improvement is observed when increasing the number of training examples.  This behavior is due to the different initial amount of prior knowledge in the two cases: when the informative MPP is used, a more accurate utility model is initially obtained and a larger amount of personal preferences from the test user are needed to sensibly update the belief about the user preferences. On the other hand, when an un-informative prior is used (no transfer learning), a less robust and less accurate initial belief is generated, and even a limited knowledge about the user preferences may sensibly change it;
\item the thin dotted straight line at 72.3\% accuracy defines the performance observed when predicting the test-user preferences by using the mass-preferences prior only ignoring individual information. 
As expected, without customizing the MPP based on the test-user's personal preferences,  a worse performance is observed. 
Furthermore, without transfer learning among users (solid curve with cross markers),  more than $10$ queries are needed to get better predictive accuracy than the one obtained when using just the MPP for predictions. 
\end{itemize}
Concluding, the observed experimental results are consistent with the intuition and clearly demonstrate that transfer learning techniques can improve route-recommendation systems in particular in the cold-start phase.

\subsection{Comparison with the alternative transfer-learning algorithm}
\label{subsec:FAVOUR_vs_ml_pl}

To finalize the evaluation of FAVOUR, its performance is compared with the alternative algorithm described in Sec.~\ref{subsec:alternative} (referred to by the acronym ALT below). The performance of ALT, estimated by applying the same LOUO-CV procedure used for FAVOUR, is shown by the dashed curve in Fig.~\ref{fig:MPPperformance}.
A statistically significant worse performance is observed for ALT w.r.t to FAVOUR. The difference is more pronounced  with few training examples, while it decreases when additional training information is used. However, even with fifteen training examples, a 2.9\% performance difference is still observed. 

To investigate the reason for the different performances, the ALT algorithm has been re-executed by using the MPPs computed by the FAVOUR approach (Sec.~\ref{sec:cold-start}) as initial user preference model. By this modification, the ALT performance becomes comparable with the FAVOUR one. No statistically significant difference is observed between the results of FAVOUR and the modified version of ALT (to avoid cluttering, the curve for the modified ALT version is not depicted in Fig.~\ref{fig:MPPperformance}, since it approximately overlaps the FAVOUR curve). 

We argue that the lower accuracy of the priors computed by the ALT algorithm is due to 
 the model selection performed during the estimation of the mixed logit models. Under normal circumstances, a model selection process is necessary to avoid overfitting. However, when a MPP is used as a start for a personalization, this model selection might actually harm the personalization process, since different users might use different variables in their selection process. However, the approach of FAVOUR of using the full variable set for the MPP does not produce a usable mixed logit model, since the parameter standard deviations would explode with the use of too many variables. To overcome this limitation, we use standard variable selection methods to get an initial model and include in it the discarded variables by assigning to them zero mean and standard deviation value similar to that of the selected parameters.  

\subsection{Comparison with the state-of-the-art}
\label{subsec:FAVOUR_vs_sta}

Our experimental evaluation is concluded by the empirical comparison between FAVOUR and the closest state-of-the-art approach introduced in~\cite{personalizedITS_2013}.
For this purpose, the approach in~\cite{personalizedITS_2013} is simulated by using a maximum-likelihood method to generate the weight vector representing the mass preferences and a vector of values quantifying the uncertainty about the estimated mass preferences. This information is used to build the initial prior for a Bayesian procedure which individualizes the weights for each test user.

The results for the benchmarking approach are summarized by the solid curve with the diamond markers in Fig.~\ref{fig:MPPperformance}. 
A superior performance of the FAVOUR algorithm is observed. FAVOUR prediction accuracy is better by $5.6$ percentage units in the case of 15 training examples. The accuracy difference increases up to the value $9.8$\%, observed when two training examples are used.  
Furthermore, with two training examples, the standard deviation of the percentage accuracy values observed for FAVOUR is $19.6$ units. In the case of 15 training examples, this value decreases down to $17.9$ percentage units. 
For the state-of-the-art approach, a larger dispersion of the accuracy values is observed: with two training examples the standard deviation is $23.6$ units, decreasing down to the value $21.1$ when 15 training examples are used. 
A more stable behavior is thus shown by FAVOUR: in the case of 15 training examples the standard deviation of the percentage accuracy values is $3.2$ percentage units lower than that of the benchmarking method, and this difference increases up to the value $4.0$ observed when two training examples are used.    
 
Since the difference between FAVOUR and the benchmarking state-of-the-art approach consists of the MPP generation, the impact of the initial (mass) preference model on the efficiency of personalized route-recommendation systems is clearly  shown. In our experiments, even 15 training examples used to customize the initial mass preference model for a new user do not enable to recover from a low-quality initial prior.

\section{Discussion}
\label{sec:conclusion}
This work improves the state-of-the-art of route-recommendations systems. In particular, 
it shows that travelers make their choices on the basis of a shared rationale and 
similar preferences. 
Therefore, route recommendations for a specific user can be improved by exploiting the preference information elicited from previous travelers. For this purpose, a Bayesian preference learning strategy (the FAVOUR algorithm) is presented, which uses a mass-preferences prior to transfer preference information among travelers. 
The performance improvement gained by the adoption of the mass-preferences prior has been experimentally demonstrated.
Furthermore, the efficiency of FAVOUR has been evaluated by a comparison with a non-Bayesian transfer-learning  technique and with a state-of-the-art route-recommendation system~\cite{personalizedITS_2013}.

While FAVOUR offers large improvements in both cases, this work leaves many avenues for future explorations. In particular, 
the adoption of more sophisticated preference models to handle the possible \emph{indifference} of the user about the presented solutions, which generates ties in route rankings~\cite{RK_ties_67}. Furthermore, ranking information may be combined with
decision maker (DM) \emph{absolute judgements}~\cite{judg_and_pref_13}, which may discard un-satisficing routes or provide hints~\cite{hints94} to the search process.
The comparison of an \emph{arbitrary} number of candidate routes (under the assumption that the pairwise comparisons are not independent) will also be evaluated.
Furthermore, the quality of the personalization of the mode-choice model based on the SP survey will need to be tested once revealed preference data becomes available within the live test system. 
Future extensions of this investigation also include the introduction of \emph{query-selection strategies}, to reduce 
 the number of comparisons asked to the DM. Promising techniques have been developed within the preference elicitation community (see, e.g.,~\cite{pe_rum_13} for a recent one) for this purpose. The reduction is achieved by the smart selection of future examples to be presented for evaluation. The selection is usually based on the current learned preference model, and therefore driven by the past feedback received from the DM. Furthermore, the predictive uncertainty of the current model is usually exploited for query selection. From this perspective, the Bayesian strategy adopted in this work is promising, as the covariance matrix of the posterior utility models provides a quantitative estimation of the predictive uncertainty.  However, during the query selection process, the routing systems may be repeatedly queried  for getting the candidate routes. The resulting computational overhead has to be evaluated, in order to achieve \emph{real-time} route recommendations.
First steps in this direction have been done in ~\cite{Chorus2013}.

The integration of real-time traffic information with customized route recommendations~\cite{VLDB15,tourism14} is another interesting line of research, since real-time traffic information influences the travelers decisions~\cite{VLDB15}. The integration of the real time traffic information into the system is easy, since the Ariadne-router already supports such functionality.  The reaction of users to repeating disturbances would be a very interesting subject for further studies since it might reveal the reaction of users to uncertain travel times.

This work gives some insights about selecting decisional features to encode real travel choices into training examples for preference learning algorithms.
However, a finer granularity to define the decisional features may be adopted
w.r.t. the one used in this work. Let us consider, e.g., the relationship between the environmental conditions and the perception of the walking time: walking ten minutes under heavy rain is clearly not as pleasant as walking ten minutes with a sunny mild spring day.

Finally, the implemented algorithms will be added to a live routing tool to present users with a selection of personalized routes. This is useful for the users, as the presented routes are closer to their personal needs, but also from a scientific standpoint, since we can collect the routes that are actually travelled as well as the choice sets handled by the users, thus producing a much larger dataset to develop and test the FAVOUR algorithm further.

\section*{Acknowledgements}
\noindent
This research was partially supported by the research projects 
emporA2 
and Crossing Borders. The projects were funded through the Climate and Energy Funds (KLIEN) of the Austrian Ministry for Transport, Innovation and Technology (BMVIT).

\bibliographystyle{IEEEtran}
\bibliography{bibtex_sources/customizedITS,bibtex_sources/ML,bibtex_sources/pref_learn,bibtex_sources/MML}

\section{Supplemental material}
\label{sec:supp}

This section contains additional information not included in the main article, due to space constraints.
In particular, Sec.~\ref{app:disp} discusses the dispersion of the FAVOUR accuracy values observed over the test sessions. In Sec.~\ref{app:pvp}, the performance of FAVOUR is measured by considering the posterior probability values associated with the correct predictions, rather than counting the number of correct predictions (predictive accuracy). This alternative performance metric measures the predictive uncertainty of FAVOUR.

\subsection{Dispersion of observed accuracy values around the mean}
\label{app:disp}

\begin{figure}%
	\centering  
	\includegraphics[width=.5\textwidth]{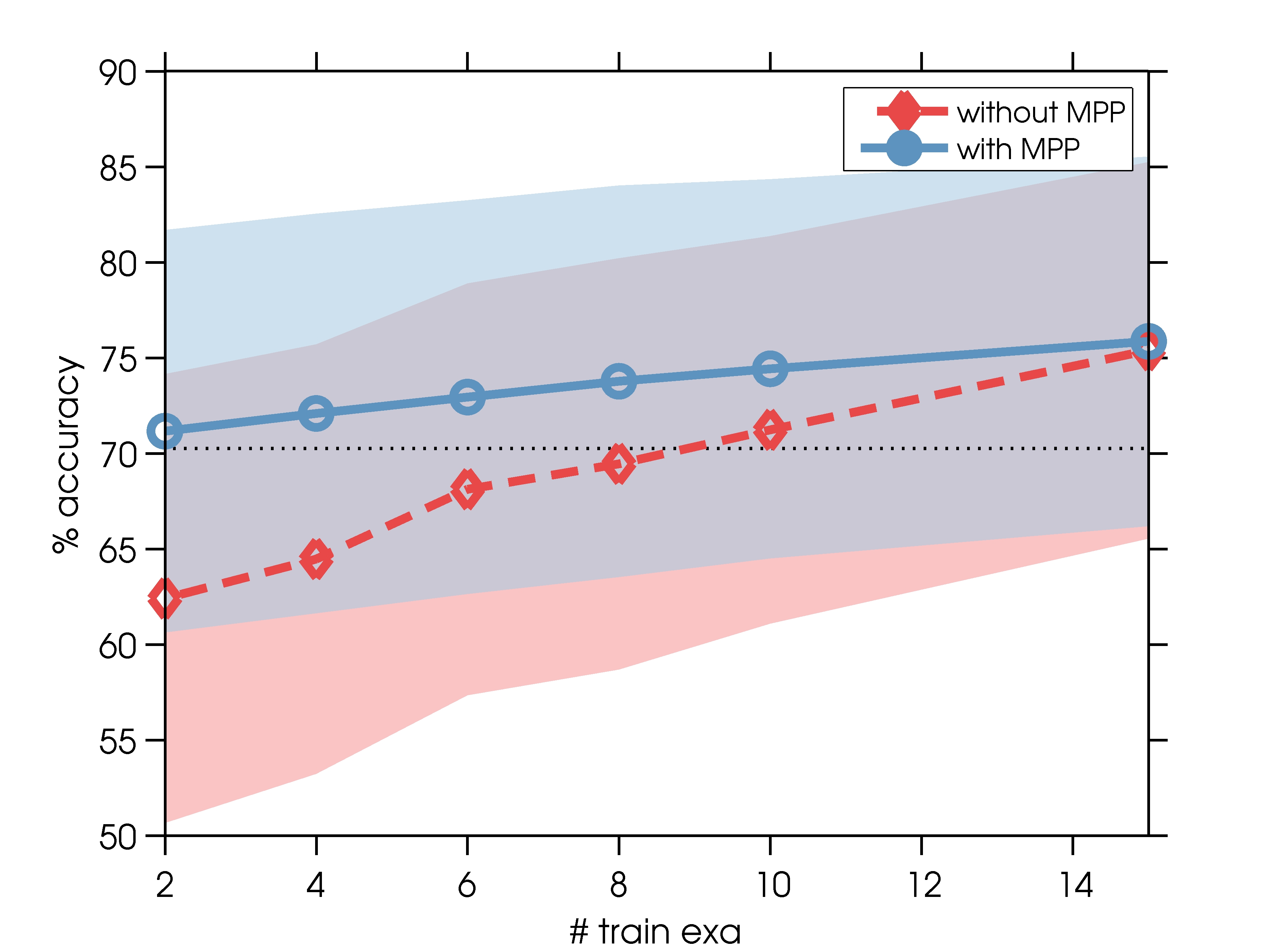}
	\caption{Learning curves observed for the FAVOUR algorithm, when the mass-preferences prior is used (solid curve with circular markers) or is not used (dashed curve with square markers). The \textit{y}-axis reports the mean accuracy returned by FAVOUR, while the \textit{x}-axis contains the number of training examples. The shaded areas denote the standard deviation of the observed accuracy values.
The thin dotted  straight line at approximately 70\% accuracy depicts the results obtained when predicting the test user preferences by using the mass-preference prior only.  Best viewed in colour.}
	\label{fig:FAVOUR_performance_acc_with_disp}
\end{figure}

The graph in Fig.~\ref{fig:FAVOUR_performance_acc_with_disp} shows the dispersion 
of the FAVOUR accuracy values observed over the test sessions. The dispersion 
around the mean accuracy is measured by the standard deviation (STD) of the accuracy values. 
When using the MPP, the STD value decreases by two percentage units, from 21\% with two training examples down to 19\% in the case of 15 examples. A similar decrease is observed when transfer learning is not adopted. The initial 23.5\% dispersion goes down to the 20\% value when increasing the number of training examples.  
Although the variability of the observed accuracy values is lower when using the MPP, the decrease in the data  dispersion is not comparable with the increase in the average accuracy. When adopting transfer learning, the average accuracy improves by 9\% and 8\% in the case of two and four training examples.    
We hypothesize that the dispersion of the accuracy values is affected by the low number of training examples used %
and by the limited size of the test set (five examples), which increases the variability of the results across the different test sets.

\subsection{Posterior probability values estimated by FAVOUR}
\label{app:pvp}

\begin{figure}%
	\centering
	\includegraphics[width=.5\textwidth]{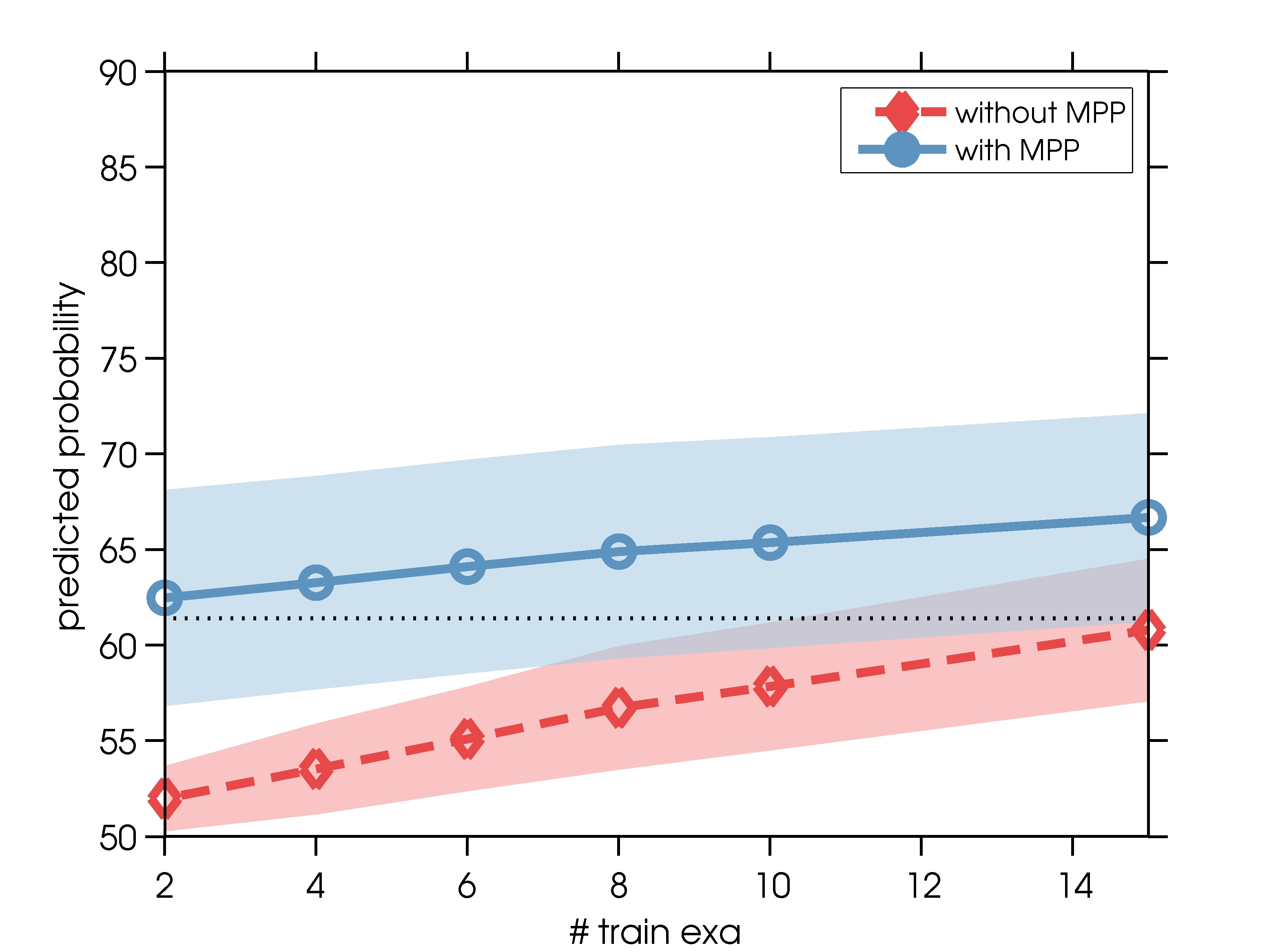}
	\caption{Learning curves observed for the FAVOUR algorithm, when the mass-preferences prior is used (solid curve with circular markers) or is not used (dashed curve with square markers). The \textit{y}-axis reports the mean probability value returned by FAVOUR, while the \textit{x}-axis contains the number of training examples. The shaded areas denote the standard deviation of the observed probability values. The thin dotted  straight line at approximately 60\%  \textit{y}-axis value depicts the results obtained when predicting the test user preferences by using the mass-preference prior only.
Best viewed in colour. 
}
	\label{fig:FAVOUR_predicted_prob_with_disp}
\end{figure}

FAVOUR adopts a probabilistic approach to learn the user preferences. The estimated 
utility model of a user is in fact a Gaussian distribution, resulting from the 
Laplace approximation of the posterior (Sec.~\ref{subsec:bl}). 
Sec.~\ref{subsec:FAVOURimpl} explains how to do inference (make predictions) with the user utility models.
In particular, FAVOUR predicts the probability
 $\mathrm{Pr} (\Br  \succ  \q |  \u(\Br), \u(\q), T)$ that the test user prefer route $\Br$ over route $\q$, given the route features $\u(\Br), \u(\q)$ and the training information $T$. A probability value strictly larger than 0.5 is interpreted as evidence that the user prefers route $\Br$. If $\mathrm{Pr} (\Br  \succ \q |  \u(\Br), \u(\q), T) \leq 0.5$, the user is expected to select route $\q$.  
This probability value estimates the predictive uncertainty of FAVOUR.
In the ideal situation, FAVOUR predicts the correct preference with probability one (certainty).
The graph in Fig.~\ref{fig:FAVOUR_predicted_prob_with_disp} shows the average probability values estimated by FAVOUR 
in the case of correct predictions. That is, the graph measures how much certain is FAVOUR when it is guessing the route preferred by the user. %
Again, the adoption of the MPP sensibly increases the performance of FAVOUR. Without transfer learning and using two or four examples only, the average confidence about the correct predictions is only 2\% units larger than the 50\% predictive certainty of random guessing. On average, 15 examples are needed to reach the predictive confidence observed when using for predictions the MPP only (thin dotted straight line), without customizing it for the test users.

\end{document}